\pdfoutput=1

\documentclass[11pt]{article}

\usepackage[final]{acl}
\usepackage{multirow}
\usepackage{booktabs} 
\usepackage{amsmath}
\usepackage{enumitem}
\usepackage{changepage}
\usepackage{array}
\usepackage[table]{colortbl}
\usepackage{dsfont}
\usepackage{amsthm}
\newtheorem{definition}{Definition}

\definecolor{CBBlue}{RGB}{0,114,178}    
\definecolor{CBOrange}{RGB}{213,94,0}  
\definecolor{CBGreen}{RGB}{0,158,115}   
\definecolor{CBRed}{RGB}{179,0,0}      
\definecolor{CBPurple}{RGB}{204,121,167}
\definecolor{CBGray}{RGB}{153,153,153}  

\usepackage{times}
\usepackage{latexsym}

\usepackage[T1]{fontenc}

\usepackage[utf8]{inputenc}

\usepackage{microtype}

\usepackage{inconsolata}

\usepackage{graphicx}

\newcommand{\claude}{\includegraphics[height=1em]{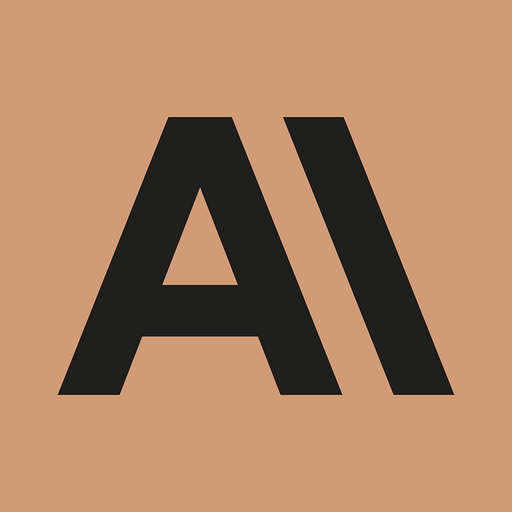}}

\title{Synergizing Unsupervised Episode Detection with LLMs\\for Large-Scale News Events}

\author{Priyanka Kargupta \quad Yunyi Zhang \quad Yizhu Jiao \quad Siru Ouyang \quad Jiawei Han\\
\quad Siebel School of Computing and Data Science\\
\quad University of Illinois at Urbana-Champaign \quad \\
\texttt{\{pk36,yzhan238,yizhuj2,siruo2,hanj\}@illinois.edu}
}

\begin{document}

\maketitle
\begin{abstract}
State-of-the-art automatic event detection struggles with interpretability and adaptability to evolving large-scale key events---unlike episodic structures, which excel in these areas. Often overlooked, episodes represent cohesive clusters of core entities performing actions at a specific time and location; a partially ordered sequence of episodes can represent a key event. This paper introduces a novel task, \textbf{episode detection}, which identifies episodes within a news corpus of key event articles. Detecting episodes poses unique challenges, as they lack explicit temporal or locational markers and cannot be merged using semantic similarity alone. While large language models (LLMs) can aid with these reasoning difficulties, they suffer with long contexts typical of news corpora. To address these challenges, we introduce \textbf{EpiMine}, an unsupervised framework that identifies a key event's candidate episodes by leveraging natural episodic partitions in articles, estimated through shifts in discriminative term combinations. These candidate episodes are more cohesive and representative of true episodes, synergizing with LLMs to better interpret and refine them into final episodes. We apply EpiMine to our three diverse, real-world event datasets annotated at the episode level, where it achieves a 59.2\% average gain across all metrics compared to baselines.
\end{abstract}

\section{Introduction}
\begin{figure}[t]
    \centering
    \includegraphics[width=0.49\textwidth]{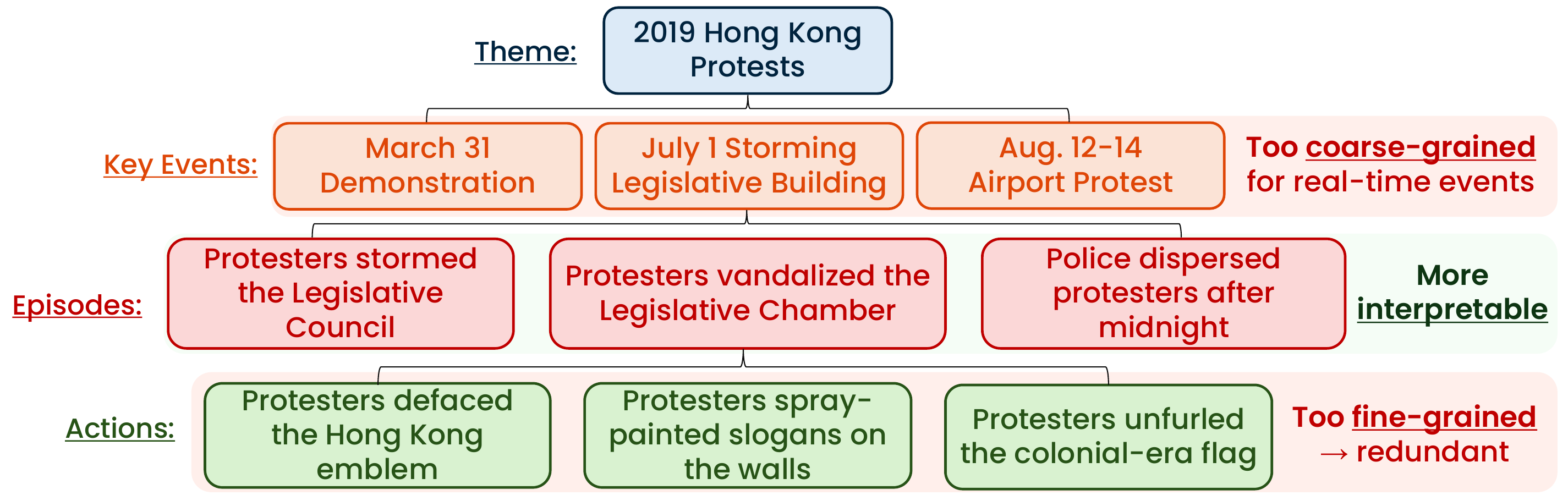}
    \caption{Example event structure hierarchy. Given a key event node's corpus, detect its episode children and their respective relevant text segments.}
    \label{fig:event_hierarchy}
\end{figure}

Given the saturation of real-time news accessible at our fingertips, reading and processing a key event's critical information has become an increasingly daunting challenge. Consequently, recent work on automatic textual event detection has attempted to integrate the manner in which humans neurologically perceive/store events into textual event detection methods. Specifically, neuroscientists studying event representations in human memory find that events are stored in a top-to-bottom hierarchy, as demonstrated in Figure \ref{fig:event_hierarchy}. The deeper the hierarchical event level, the more fine-grained its corresponding text granularity \cite{zhang2022unsupervised}: we consider a theme as corpus-level (all articles discussing the 2019 Hong Kong Protests), key event as document-level (an article typically discusses a full one to two day key event), episode as segment-level, and atomic action as sentence or phrase-level.

Furthermore, neurological research \cite{baldassano2017discovering, khemlani2015episodes} indicates that events are encoded into memory as \textit{episodic structures}. Representing events as discrete episodes helps us piece together a \textit{coherent and concise narrative} by focusing on \textit{meaningful} clusters of actions, reactions, and developments, rather than examining each in isolation or as a whole. Despite its strengths, \textbf{existing automatic event extraction works fail to consider the episode-level}. 

\par For instance, key event detection focuses on identifying ``a set of thematically coherent documents'' for each key event \cite{zhang2022unsupervised, liu2023key}, but manually parsing large clusters of articles is inefficient and lacks interpretability. Timeline summarization \cite{steen-markert-2019-abstractive, li-etal-2021-timeline, ghalandari2020examining, chen2023follow} addresses this by providing dates and compact summaries, yet it suits historical themes better than evolving key events that require finer granularity. Event chain mining \cite{jiao2023unsupervised} takes a fine-granularity approach by identifying temporally ordered atomic actions, but its phrase-level granularity is often too fine and practically redundant for large-scale events (e.g., in Figure \ref{fig:event_hierarchy}, the actions all describe the same episode). To bridge this gap, we propose the novel task of \textbf{episode detection} to pave the way for a more effective event representation.

\par Episode detection aims to detect episodes from a news corpus containing key event articles. An episode can be described as a cohesive cluster of potentially diverse subjects performing actions at a certain time and location, occurring as part of a larger sequence of episodes under a specific key event. We introduce \textbf{EpiMine}, which detects meaningful \underline{episodic events} and their \underline{corresponding text segments} in a large key event corpus, all without any level of human supervision or labeled training data. EpiMine consists of: (1) episode indicative term mining, (2) episode partitioning, (3) LLM-enhanced episode estimation, and (4) episode-segment classification. Collectively, they tackle the unique challenges of episode detection, detailed below:

\par{\textbf{Challenge 1: Episodes are not timestamped.}}
Key event detection partitions a thematic corpus into document-level clusters by heavily relying on explicit temporal features, like publication dates, being associated with the key event articles \cite{zhang2022unsupervised}. However, this assumption fails at the episode-level, where there is no guarantee to have a distinct timestamp associated with each text segment that discusses a new episode. Fortunately, we can take advantage of the idea that journalists naturally partition news articles by sequentially discussing distinct episodes:
\begin{adjustwidth}{0.5em}{0em} \textbf{Example:} An article likely completes its discussion of the episode A, \textit{protesters storming the Legislative Council}, before episode B, ``\textit{protesters vandalized the Legislative Chamber}'' (Figure \ref{fig:partition}).
\end{adjustwidth}
\noindent Hence, to partition articles into distinct episode segments, EpiMine must identify whether two consecutive segments are discussing the same or different episodes--- bringing us to our next challenge.

\par{\textbf{Challenge 2: Episodes contain semantically diverse actions.}} Each episode features a \textit{set of unique atomic actions}, which can help determine if two segments discuss the same episode. However, for clustering actions, existing methods \cite{jiao2023unsupervised} rely heavily on semantic similarity. This is not realistic for episode-segment clustering:
\begin{adjustwidth}{0.5em}{0em} \textbf{Example:} ``\textit{protesters spray-painted slogans}'' and ``\textit{they unfurled the colonial-era flag}'' will fall under the same episode, but are semantically different and unlikely to be clustered.
\end{adjustwidth}
\noindent Alternatively, we can identify notable (salient) terms unique to the same episode (episode A: ``barriers'' and ``shoved''; episode B: ``\textit{defaced}'' and ``\textit{walls}''), by exploiting corpus-level signals. For example, if ``\textit{defaced}'' and ``\textit{walls}'' are frequently mentioned together across the corpus (or their respective synonyms) and not with other terms, then they are a \textit{discriminative co-occurrence}. When terms between two segments discriminatively co-occur, this indicates the same episode is being discussed. Conversely, if a sufficient shift in term combinations occurs, then a different episode is being discussed.

\par{\textbf{Challenge 3: Articles often do not feature all episodes.}} Real-time news reporting often provides an incomplete coverage of multi-day events, with individual articles potentially omitting or partially addressing key episodes. Consequently, while LLMs could assist with the first two challenges, requiring multiple articles hinders their use given their long context limitations \cite{li-etal-2024-loogle, liu-etal-2024-lost}. To address this challenge, EpiMine seeks to select a minimal set of articles that maximizes both the quantity and quality of event episodes. It then merges any article partitions across these articles which likely discuss the same episode and synergizes with an LLM to provide a more fluent interpretation of the candidate episodes, accounting for the episode's core entity, actions, object, location, and time period. This allows EpiMine to finally map the remaining non-salient article segments to these episodes, pruning any candidates which are not sufficiently supported by the remaining articles. We summarize our core contributions:
\leftmargini=12pt
\begin{itemize}
\itemsep0.1em 
\item \textbf{Episode detection}: \textit{novel} task to detect episodes \& their segments from a key-event corpus.
\item \textbf{EpiMine}, an unsupervised episode detection method which introduces discriminative term co-occurrence and episode partitioning.
\item \textbf{Three novel datasets}, reflecting a diverse set of real-world themes and thirty global key events (no key event corpus exists for this task).
\item EpiMine outperforms all baselines by, on average, a \textbf{59.2\% increase across all metrics}.
\end{itemize}
\noindent
\textbf{Reproducibility:} We provide our dataset and source code\footnote{\url{https://github.com/pkargupta/epimine}} to facilitate further studies.

\section{Related Works}
\subsection{Event Extraction}
Event extraction has been widely studied, focusing on event detection \cite{liu-etal-2018-jointly, DBLP:conf/emnlp/DuC20, DBLP:conf/naacl/LiJH21, lu-etal-2021-text2event, qi-etal-2022-capturing, DBLP:conf/emnlp/JiaoLXZJH22}, event relation extraction \cite{DBLP:conf/conll/HanHYGWP19, DBLP:conf/emnlp/WangCZR20, DBLP:conf/aaai/AhmadPC21}, and salient event identification \cite{DBLP:conf/emnlp/LiuXMH18, DBLP:conf/coling/JindalDR20, DBLP:conf/emnlp/WilmotK21}. Recent work has also addressed event process understanding \cite{DBLP:conf/emnlp/ZhangCWSR20, DBLP:conf/conll/ChenZWR20}, though these often rely on expensive expert annotations. Some studies have introduced unsupervised methods to address annotation challenges \cite{DBLP:conf/emnlp/WeberSBC18, DBLP:conf/emnlp/LiZLCJMCV20}. Some overlapping work exists in topic discovery, where \cite{yoon2023unsupervised} proposes unsupervised stream-based story discovery--- computing article embeddings based on their shared temporal themes. Recently, large language models have demonstrated powerful general and event extraction-specific reasoning abilities \cite{pai2024survey, gao2024eventrl}.
\par However, traditional and LLM-driven methods either, (1) focus on phrase/sentence-level events (analogous to actions in Figure \ref{fig:event_hierarchy}), or (2) require human-curated event ontologies, often overlooking interpretable, yet meaningful granularities and open-domain texts, which go beyond pre-defined event types. While unsupervised granular event extraction has been explored \cite{zhang2022unsupervised, jiao2023unsupervised} at the document and phrase-level, episode detection is a more interpretable granularity that remains a largely unexplored, yet vital area.

\subsection{Timeline Summarization}
Timeline summarization (TLS) identifies key dates and concise descriptions for major events. Early methods were extractive, focusing on ranking events for thematic timelines \cite{nguyen2014ranking} or using submodular frameworks to model temporal dimensions \cite{martschat2018temporally}. Abstractive methods later emerged, such as sentence clustering and multi-sentence compression \cite{steen-markert-2019-abstractive}. More recent approaches are graph-based, such as event-graph representations for salient sub-graph compression \cite{li-etal-2021-timeline} and heterogeneous GATs for redundancy reduction \cite{you2022joint}. While they effectively summarize key events as high-level timelines, they focus on historical themes. Episode-level timelines for ongoing news remain underexplored.

\begin{figure*}[ht!]
    \centering
    \includegraphics[width=1.0\textwidth]{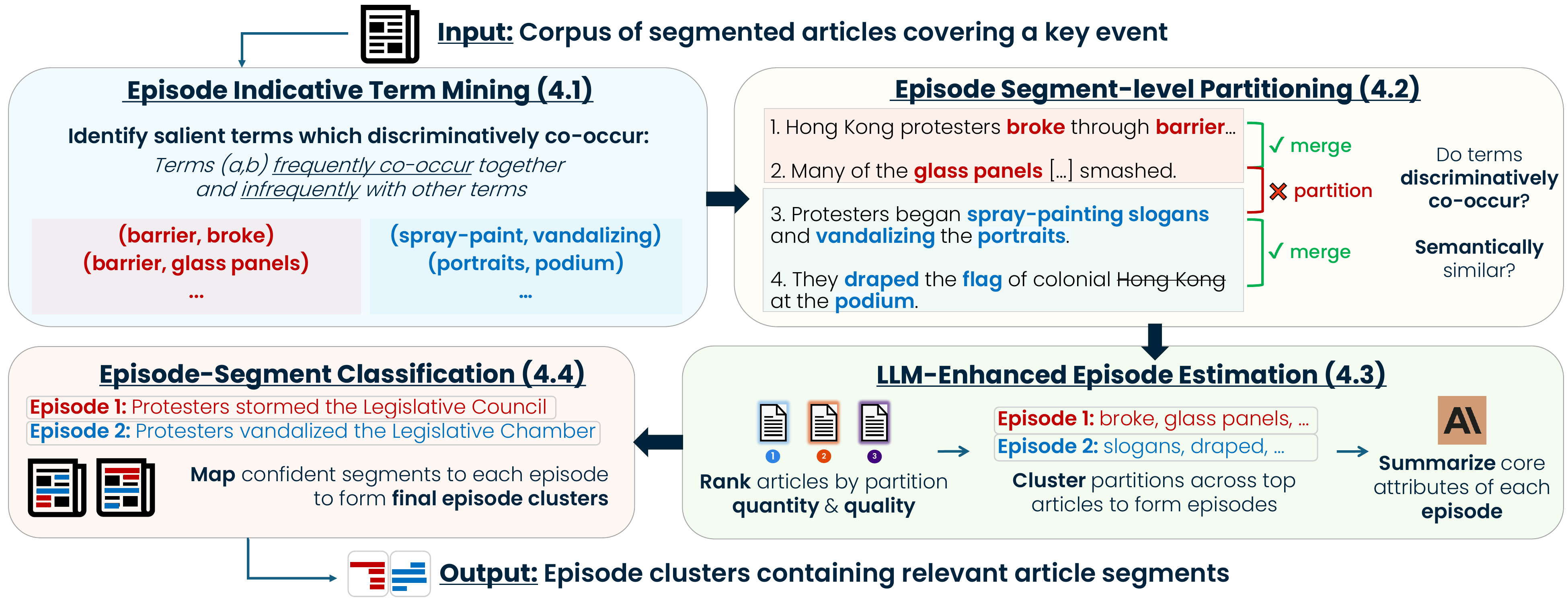}
    \caption{
    \label{fig:framework} We detail the overall framework of EpiMine.}
\end{figure*}

\section{Preliminaries}
\subsection{Problem Definition}
\label{subsec:problem_definition}
\begin{definition}[\textbf{Episode}] An \underline{episode} $E_i$ is one of a partially ordered sequence of subevents, $\{E_1, 
\ldots, E_i, \ldots E_k\}$, of a key (major) event $E$, where typically $2 \leq k \leq 20$, and $E_i$ does not overlap with $E_j$ if $i \neq j$. Actions in the episode $E_i$ can be either semantically similar or diverse, but typically have relatively tight time, location, and/or thematic (entities, actions, objects) proximity.\end{definition}
EpiMine aims to extract episodes from a news corpus, where an episode occurs as a significant component of a larger group of episodes that fall under a specific key event. For instance, in Figure \ref{fig:event_hierarchy}, without knowing Episode \#1, ``Protesters stormed the Legislative Council Complex'', readers would not fully understand Episode \#3, ``Police dispersed protesters''. Hence, episodes help us understand the overall key event and are especially useful for events that are currently evolving, where finer context is required for sufficiently understanding them.
\begin{definition}[\textbf{Episode Detection}] Given a corpus $\mathcal{D}$ about one key event, where each document $d \in \mathcal{D}$ is a news article, the task is to obtain a set of text segment clusters $\mathcal{E} = \{E_1, E_2, \ldots, E_k\}$. Each episode cluster $E_i \subset \mathcal{S} = \{s_1^1, s_2^1, \ldots, s_{|d|}^{|\mathcal{D}|}\}$, where $\mathcal{S}$ contains all the text segments identified in each document $d \in \mathcal{D}$, and every two clusters do not have overlapping text segments (i.e., $E_i \cap E_j = \emptyset$ for $i \neq j$).
\end{definition}
It is important to note that $k$, the number of episodes, is not known in advance and oftentimes, a news article segment may discuss either episodes of a different key event (e.g., an episode with similar aspects that occurred in a different historical key event) or multiple episodes of the current key event. Nonetheless, our goal is to detect the most relevant episodes to the current key event at hand and consequently mine the most distinctive text segments for each of these (hence our constraint of $E_i \cap E_j = \emptyset$ for $i \neq j$).

\section{Methodology}
To tackle episode detection, we propose a novel unsupervised framework, EpiMine. As shown in Figure \ref{fig:framework}, EpiMine consists of the following four core components: (1) \textbf{episode indicative term mining}, which identifies combinations of salient terms likely to discriminatively co-occur \textit{within} an episode and not \textit{across} episodes; (2) \textbf{episode partitioning}, which partitions each article into approximate isolated episodes based on consecutive shifts in the term co-occurrence distribution, (3) \textbf{LLM-enhanced candidate episode estimation}, which clusters the top partitions into candidate episodes and utilizes LLM-based reasoning to produce fluent and meaningful episodes, and (4) \textbf{episode-segment classification}, which maps confident segments to their respective episode clusters.

\subsection{Episode Indicative Term Mining}
\label{sec:term_mining}
\begin{figure}[ht]
    \centering
    \includegraphics[width=0.45\textwidth]{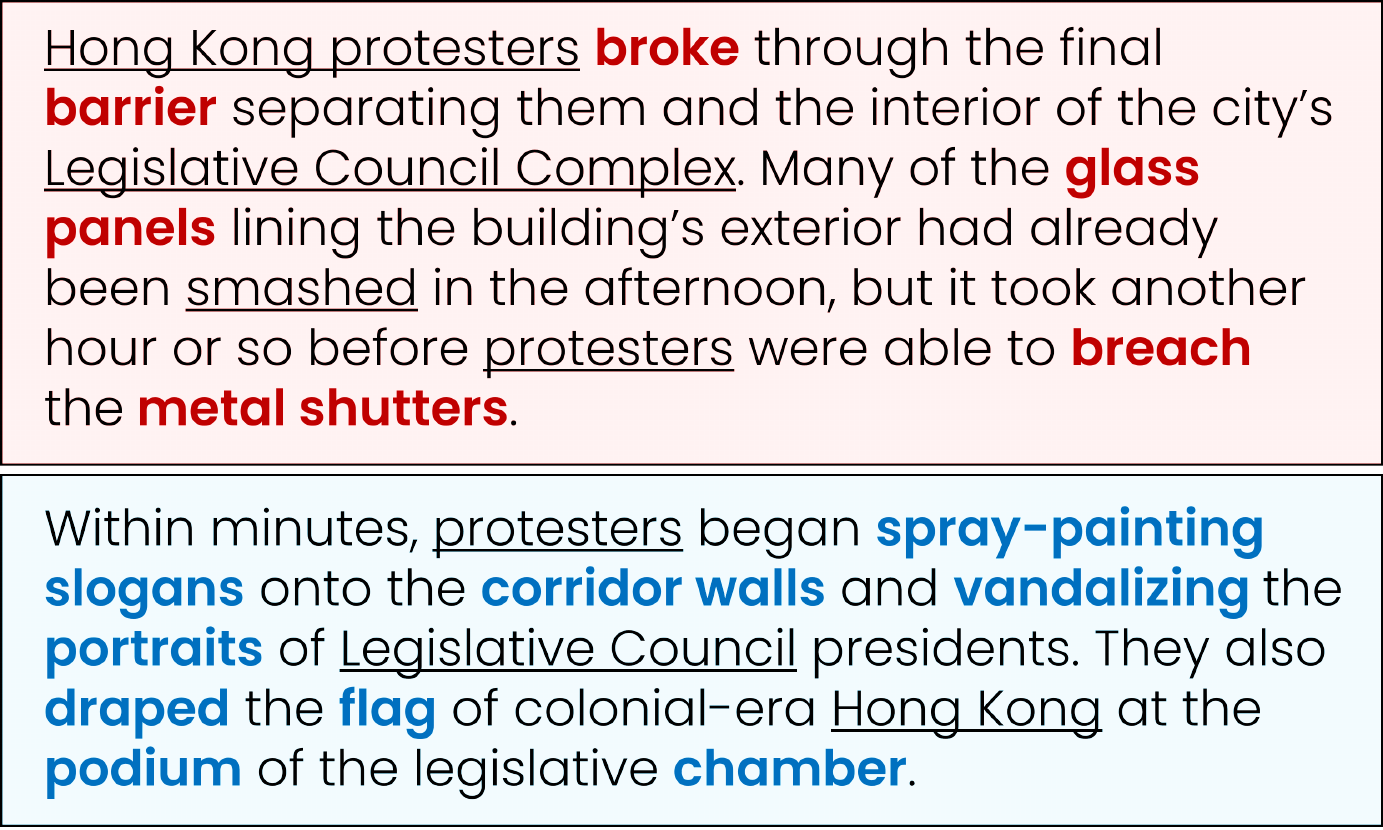}
    \caption{Natural partition between two episodes in a key event article. An episode's discriminative terms are \textbf{bolded}; salient non-discriminative terms are \underline{underlined}.}
    \label{fig:partition}
\end{figure}
\par Lacking supervision, our goal is to identify \textit{potential} candidates for episodes. Episodes are often described in relation to each other and usually lack timestamps or locations consistently mentioned within their segments. For example, the phrase ``police dispersed protesters'' may not have a precise timestamp because it is a \textit{response} to ``protesters stormed the Legislative Council Complex,'' and some journalists may consider the implicit ordering adequate. 
Additionally, the same episode can be described using different entities and actions--- journalists may \textit{report different perspectives}. For example, both ``protesters shoved against the barricades'' and ``the police used pepper-spray on the protesters'' describe the episode ``protesters stormed the Legislative Council Complex''. However, they are semantically different, focused on different core entities and actions. Thus, we cannot depend on a consistent subject-action-object triple or an explicit time/location mapped to each episode in the article.

\par To circumvent this challenge, we exploit the idea that journalists naturally partition news articles according to episodes, forming \underline{episode fragments}. For example, as shown in Figure \ref{fig:partition}, an article will likely complete its discussion of episode \#1, ``Protesters stormed the Legislative Council Complex'' (red), before fully shifting to discussing episode \#2, ``protesters vandalized the Legislative Chamber'' (blue). Across these episode fragments, certain \textit{salient} terms are featured (e.g., protesters, legislative, vandalizing, podium). We adapt the idea of event salience from \cite{jiao2023unsupervised} specifically for the task of episode detection.

\begin{definition}[\textbf{Salience}]
    A term is \underline{salient} if it is (1) \textbf{distinct and significant} to understanding a given key event, as well as (2) \textbf{frequently} found in a key event's segments and \textbf{infrequently} in other background/general articles.
\end{definition}

Thus, we identify a set of salient terms for each segment within the corpus (salience score details in Appendix \ref{sec:salient}).

\par{\textbf{Discriminative Co-occurrence.}} In Figure \ref{fig:partition}, we can see that the first episode fragment, and Episode \#1 in general, features a combination of similar terms, such as ``protesters'', ``barrier'', and ``breach''. Likewise, the second episode may include a combination of terms similar to ``protesters'', ``spray-painting'', and/or ``flag''. We note that despite some journalists choosing to only describe the protesters spray-painting, while others focus on the protesters draping the colonial-era flag, we must be able to recognize that their respective salient terms are \textit{likely to co-occur within the same episode}. 

\par However, we make a \textbf{novel distinction} between a co-occurrence and a \textit{\textbf{discriminative}} co-occurrence. Salient terms $a$ and $b$ (e.g., ``protesters'' and ``spray-painting'') may often co-occur within an episode. However, if $a$ also frequently co-occurs with many other terms in various episodes (``protesters broke''), $a$ and its co-occurrences are less useful for distinguishing episodes. Thus, $(a,b)$ is not a discriminative co-occurrence.

\begin{definition}[\textbf{Discriminative Co-occurrence}]
A pair of terms $(a,b)$ \underline{discriminatively co-occur} if (1) they frequently appear together in episode $E_i$, and (2) neither $a$ nor $b$ appear \textit{as frequently} with other terms $w$ in other episodes $E_{\notin i}$.
\end{definition}
We compute the discriminative occurrence $d$ between salient term pair $(a,b)$ using the following:
\begin{equation}
\small
    \label{tag:discriminative}
    \begin{split}
        \textbf{d}(a,b) = &\log\left(\frac{freq(a,b)}{\max(\bar{f_{a}}, \bar{f_{b}})}\right) \times \log\left(\frac{|T|}{max(|F_{a}|, |F_{b}|)}\right), \\
        \text{where } &\bar{f_{a}} = \frac{1}{|T|}\sum\limits_{\forall w_i \in T} freq(a, w_i) \text{,  and} \\
        & F_{a} = \{freq(a, w_i) > 1 \text{  } \forall \text{  } w_i \in T\}
    \end{split}
\end{equation}

The first $\log$ term ensures that the pair's co-occurrence ($freq(a,b)$) is \textbf{statistically significant} ($\geq$ the max of $a$ and $b$'s mean vocabulary-wide co-occurrence respectively). The second $\log$ term ensures the pair is a \textbf{discriminative match}, penalizing cases where $a$ or $b$ frequently co-occurs with a large portion of the salient term set $T$. For example, co-occurrences with ``protesters'' are not discriminative because ``protesters'' is a core entity in all episodes and thus frequently co-occurs with many terms in $T$. In contrast, (``slogans'', ``flags'') is a discriminative co-occurrence since both terms frequently appear together in segments discussing episode \#2 and rarely co-occur with other terms $w_i \in T$. If $a$ and $b$ are the same term or close synonyms (determined by statistically significant semantic similarity), they have maximum co-occurrence. By leveraging multiple articles in a large key event corpus, we have sufficient statistical support to ensure our output reflects the average realistic reporting of the key event and its episodes.

\subsection{Episode Partitioning}
\label{sec:episode_partitioning}

\par With the ability to identify discriminative co-occurrences, we can use a key transitive property to resolve episode co-references within and across articles, where \textit{not all combinations of an episode's discriminative terms explicitly co-occur}:
\begin{quote}
    If $(a,b)$ and $(b,c)$ are both discriminative co-occurrences, then $(a,c)$ is \textit{also likely} to be a discriminative co-occurrence.
\end{quote} 
To illustrate this, we have the following text segment excerpts of a news article (the salient and discriminative terms are \textit{italicized}):
\begin{enumerate}
\itemsep-0.25em
    \item Protesters \textit{defaced} the Hong Kong \textit{emblem}, \textit{spray-painted slogans}, and \textit{unfurled} the \textit{flag}.
    \item The \textit{portrait} of LegCo president was \textit{defaced}.
    \item A \textit{slogan} on the \textit{wall} reads: ``The government forced us to revolt''.
    \item \textit{Police} said at least 13 people had been \textit{arrested} on \textit{suspicion} of involvement in the pro-democracy protest.
\end{enumerate}

\par We can naturally see that segments 1-3 all discuss the ``protesters vandalized the Legislative Chamber'' episode, while segment 4 discusses the ``police dispersed protesters'' episode. We can systematically replicate this partitioning process by considering the discriminative co-occurrence score between all pairwise combinations of terms from segments ($i-1$) and ($i$). If the average discriminative co-occurrence \textit{and} static semantic similarity between each term $a$ from ($i-1$) and $b$ from ($i$) is statistically significant ($\geq \mu_{d} - \sigma_{d}$) for that specific article $d$ (e.g., notably (slogans, defaced) for segments 1-3), we hypothesize that the \textbf{same episode} is being discussed and \textit{merge} them into one episode fragment. If not (e.g., (slogans, arrested) for segments 3-4), this indicates that a \textbf{different episode} is being discussed, and we \textit{partition} them into two episode fragments. Further implementation details are provided in Appendix \ref{sec:partitioning_details}.

\subsection{LLM-Enhanced Episode Estimation}
\label{sec:candidate_estimation}

\par LLMs demonstrate strong event-specific reasoning at the phrase or sentence level \citep{pai2024survey, gao2024eventrl}, but they struggle with understanding long contexts \citep{li-etal-2024-loogle, liu-etal-2024-lost}. This limitation hampers their ability to process all episode fragments for detecting episodes. Additionally, noisy retrieval significantly affects reasoning performance \citep{shen-etal-2024-assessing}. To address these challenges, we propose a synergistic approach that enhances in-context episode reasoning by reducing the number of required fragments while improving their cohesiveness and quality. We first identify the set of articles that maximizes the \textit{quantity and quality} of potential episodes, where each article is ranked by multiplying two metrics:
\begin{enumerate}
    \itemsep-0.5em
    \item \textit{Quality of episode fragments}: A top article should primarily consist of episode fragments containing salient terms that discriminatively co-occur. This reduces the rank of general fragments which summarize/analyze the event. We average each episode fragment's mean inner-discriminative co-occurrence (across all pairwise combinations of its salient terms).
    \item \textit{Quantity of episode fragments}: A top article should ideally contain all ground-truth episodes. Therefore, we take the $\log$ of the number of episode fragments in the article.
\end{enumerate}

\par After ranking all articles, we select the top $\delta$\% and resolve potential co-references to the same episode across these top articles. We apply agglomerative clustering \cite{murtagh2012algorithms} to the top episode fragments using a pre-computed distance matrix. The distance between two fragments (inversed) is calculated using the same discriminative and static semantic similarity score used in Section \ref{sec:episode_partitioning}). Clusters with a statistically insignificant number of episode fragments are pruned.
\par Finally, we provide episode fragment clusters as a more interpretable context for the LLM to resolve two challenges: (1) missing time and location stamps in fragments, and (2) semantic inconsistencies within clusters. The LLM summarizes each cluster by identifying its core attributes--- entities, actions, objects, location, and time. It then outputs the \textit{episode attributes}, \textit{relevant keywords} for extraction, and the top \textit{extracted text segments} (prompt \& example in Appendix \ref{sec:prompt}).

\subsection{Episode-Segment Classification} 
\label{sec:classification}
\par With these core summaries of the episode clusters, we obtain a generalized description of each candidate episode. For each candidate, we encode its LLM-based core attributes and extracted segments to compute a simple episode representation. Specifically, following extremely weakly supervised text classification works \cite{wang-etal-2021-x, kargupta-etal-2023-megclass}, we take the harmonic mean of these representations---as the latter extracted segments are likely not as significant as the earlier extractions and core attributes. We similarly encode all input \textit{segments} with the same encoder. We use these to assign an episode and confidence score to each encoded input segment.

\paragraph{\textbf{Episode-Segment Confidence Estimation.}} 

\par Directly mapping a text segment to its top episode based on cosine similarity risks misclassifying episode-irrelevant segments or those discussing multiple episodes (e.g., a journalist's summary). To avoid classifying such segments and ensure non-overlapping episode clusters (as discussed in Section \ref{subsec:problem_definition}), we must determine the confidence of a segment discussing a single episode.

\par We compute segment $s_i$'s cosine similarity to its top two episodes ($e_i^0$ and $e_i^1$). A larger gap ($e_i^0 - e_i^1$) reflects greater confidence in classifying $s_i$ to $e_i^0$. Each gap is normalized by the sum of all segment-episode gaps across the corpus, ensuring confidence is relative to the key event:
\begin{flalign}
    \label{tag:episode_confidence}
    &s_{i, \text{confidence}} = \frac{e_i^0 - e_i^1}{\sum_{l=1}^{|\mathcal{S}|} (e^0_l - e^1_l)}
\end{flalign}

\par Segments with statistically significant confidence in their top episode are assigned to their respective episode clusters $E_i$. Episodes with no assigned segments are pruned, yielding the \textbf{final detected episodes and clusters, $\mathcal{E}$}.

\section{Experiments}
For implementing \textbf{EpiMine}, we use the following hyperparameters across all datasets: $\delta=25\%$, $sim\_thresh=0.75$. We also use Claude-2.1 as our base LLM (\claude). All other hyperparameters are set to their respective default values. We provide all experimental settings in Appendix \ref{sec:settings}.

\begin{table}[h]
\centering\setlength{\tabcolsep}{6.5pt}
\footnotesize
  \caption{Statistics of our collected datasets. The numbers are averaged per key event.}
  \label{table:dataset_stats}
  \begin{tabular}{lccc}
    \toprule
    \textbf{Theme} &\textbf{\# docs} & \textbf{\# episodes} & \textbf{\# segments} \\
    \midrule
    \textbf{Terrorism/Attacks} & 32.2 & 5.9 & 290.3\\
    \textbf{Natural Disasters} & 36.2 & 7.4 & 324.6\\
    \textbf{Political Events} & 70.2 & 7.5 & 667.7\\
  \bottomrule
  \vspace{-0.5cm}
\end{tabular}
\end{table}

\begin{table*}[ht!]
\center \small
\caption{Results averaged across each theme, including the mean \# of episodes that EpiMine identifies per theme (in parenthesis). Results are computed on each key event corpus using the top-5 documents for each detected episode. Due to variance in LLM generation, we run it 10 times and report the average of each measure. We scale each value by 100. Bold values denote the top method; second-best method is \underline{underlined}.}
\label{tab:results}
\begin{tabular}{l|ccc|ccc|ccc}
\toprule
& \multicolumn{3}{c}{\textbf{Terrorism} (5.36 eps)} & \multicolumn{3}{c}{\textbf{Natural Disasters} (7.4 eps)} & \multicolumn{3}{c}{\textbf{Politics} (7.5 eps)} \\
\cmidrule(lr){2-4} \cmidrule(lr){5-7} \cmidrule(lr){8-10}
Methods & \textit{5-prec} & \textit{5-recall} & \textit{5-F1} & \textit{5-prec} & \textit{5-recall} & \textit{5-F1} & \textit{5-prec} & \textit{5-recall} & \textit{5-F1} \\
\midrule
EMiner &  8.64 & 0.25 & 0.48 & 10.37 & 0.19 & 0.37 & 8.66 & 0.16 & 0.32 \\
\midrule
K-means & 21.23 & 21.23 & 21.23 & 27.85 & \textbf{28.47} & \underline{28.14} & 16.04 & 16.04 & 16.04 \\
K-means + \claude & 28.58 & 14.04 & 18.26 & 37.40 & 16.58 & 22.00 & 27.18 & 17.36 & 18.25 \\
\midrule
EvMine & 23.03 & 15.02 & 17.45 & 28.15 & 8.02 & 12.25 & 5.36 & 4.00 & 4.58 \\
EvMine + \claude & 37.88 & 15.70 & 21.33 & 43.56 & 13.22 & 19.40 & 32.73 & 12.98 & 17.28 \\
\addlinespace
\hline
\addlinespace
\textbf{EpiMine (\claude)} & \textbf{71.21} & \underline{22.07} & \underline{32.43} & \textbf{70.98} & \underline{28.46} & \textbf{34.53} & \textbf{62.67} & 21.54 & \textbf{29.23} \\
 - No Confidence (\claude) & \underline{61.97} & \textbf{30.19} & \textbf{38.45} & \underline{43.66} & 20.78 & 27.76 & \underline{60.29} & \textbf{27.73} & \underline{24.77} \\
 - No LLM & 37.73 & 21.62 & 24.77 & 37.19 & 14.78 & 17.52 & 30.64 & \underline{23.51} & 19.06 \\
\bottomrule
\end{tabular}
\end{table*}

\subsection{Datasets}
\label{keyeventlist}
\par We conduct our experiments on three novel thematic, real-world news corpora selected from Wikipedia\footnote{https://en.wikipedia.org/wiki/} over the last decade. For each theme, we manually collect approximately $10$ key events composed of multiple articles and ensure that distinct \textit{episodes} exist in each:
\begin{itemize}
\small
    \item \textbf{Terrorism and attacks:} \textit{2021 Atlanta spa shootings; 2014 Montgomery County Shootings; 2021 Indianapolis FedEx shooting; 2022 Cincinnati FBI field office attack; 2019 Jersey City shooting; 2019 Naval Air Station Pensacola shooting; 2022 Greenwood Park Mall shooting, 2018 Capital Gazette shooting; 2021 Collierville Kroger shooting; 2019 Kyoto Animation arson attack}
    \item \textbf{Natural disasters:} \textit{2023 Tornado outbreak sequence; 2023 Hawaii Wildfires; 2021 Western Kentucky tornado; 2017 Mocoa landslide; 2010 Haiti earthquake; 2021 Henan floods; 2019 Nyonoksa radiation accident; 2022 NA winter storm; 2011 Fukushima nuclear accident}
    \item \textbf{Political events:} \textit{2020 Kyrgyz Revolution, 2019 Storming of the Hong Kong Legislative Council Complex; 2019 Siege of the Hong Kong Polytechnic University; 2017 Zimbabwean coup; 2018 Italian government formation; 2021 January 6 U.S. Capitol attack; 2018 Thai Cave Rescue Operation; 2018 Armenian Revolution; 2017 Lebanon–Saudi Arabia dispute; 2013 Tunisian political crisis}
\end{itemize}
\normalsize
\par The articles are obtained from the Wikipage references of each key event--- filtered with constraints in time, language, and relevance. Furthermore, each article is segmented to match our setting (Appendix \ref{sec:segment}), and each segment is automatically annotated (Appendix \ref{sec:annotation_prompt}) with either its corresponding episode ID or with a multiple/no episode tag (`M' or `X'). Further details on the criteria/process, each theme, and corresponding key events are in Appendices \ref{sec:keyeventinfo} and \ref{sec:annotation_prompt}. We also conduct a human-automatic agreement analysis for segment-episode annotation, which shows substantial agreement with good reliability (Appendix \ref{sec:keyeventinfo}).

\subsection{Baselines}
We compare against the following methods using the evaluation metrics specified in Appendix \ref{sec:metrics}: \textbf{(1)} \textbf{K-means}~\cite{likas2003global}: \underline{given} the \# of ground-truth episodes, it clusters segments using ST ~\cite{reimers2019sentence} embeddings; \textbf{(2) EvMine}~\cite{zhang2022unsupervised}: a document-level unsupervised key event detection method adapted to segment level for episode detection; \textbf{(3) EMiner}~\cite{jiao2023unsupervised}: unsupervised event chain miner that clusters atomic actions, adapted to episodes; \textbf{(4) No Confidence}: an ablation that uses max cosine-similarity instead of confidence from Equation \ref{tag:episode_confidence}; \textbf{(5) No LLM}: an ablation that uses estimated episode clusters from Section \ref{sec:candidate_estimation} to compute our episode representations directly. We also integrate \claude \space into K-means and EvMine using our same prompt (Appendix \ref{sec:prompt}). All baseline and ablation details are in Appendix \ref{sec:baselines}.

\begin{table*}[h!]
\centering\setlength{\tabcolsep}{6.5pt}
\small
  \caption{Ablation studies conducted on top 25\% of article episode clusters (Section \ref{sec:candidate_estimation}).}
  \begin{tabular}{l|ccc|ccc|ccc}
    \toprule
    & \multicolumn{3}{c}{\textbf{Terrorism}} & \multicolumn{3}{c}{\textbf{Natural Disasters}} & \multicolumn{3}{c}{\textbf{Politics}} \\
    \cmidrule(lr){2-4} \cmidrule(lr){5-7} \cmidrule(lr){8-10}
    Ablations & \textit{5-prec} & \textit{5-recall} & \textit{5-F1} & \textit{5-prec} & \textit{5-recall} & \textit{5-F1} & \textit{5-prec} & \textit{5-recall} & \textit{5-F1} \\
    \midrule
    \textbf{EpiMine-Top} & \textbf{0.2292} & \textbf{0.2435} & \textbf{0.2144} & \textbf{0.3817} & \textbf{0.2232} & \textbf{0.2450} & 0.1051$^\dagger$ & \textbf{0.2233} & 0.1201$^\dagger$ \\
    \textbf{TF-IDF} & 0.0985 & 0.1403 & 0.1059 & 0.3284$^\dagger$ & 0.1919$^\dagger$ & 0.2221$^\dagger$ & 0.0907 & 0.1908 & 0.0916 \\
    \textbf{No DC} & 0.1968$^\dagger$ & 0.1752$^\dagger$ & 0.1707$^\dagger$ & 0.2520 & 0.1546 & 0.1785 & \textbf{0.1126 }& 0.2108$^\dagger$ & \textbf{0.1299}\\
  \bottomrule
\end{tabular}
\label{tab:ablations}
\end{table*}

\begin{table}[h]
    \small
    \centering
    \caption{Compares top-5 salient terms which (1) have the highest cosine-sim (CS) and (2) discriminative co-occurrence (DC), with the given keyword.}
    \begin{tabular}{p{1.1cm}|p{2.5cm}|p{2.5cm}}
        \hline
        \textbf{Keyword} & \textbf{CS} & \textbf{DC} \\ 
        \hline 
        broke & stormed, ransacked, dashed, occupied, rushed & glass, doors, metal, building, teargas \\
        \hline
        slogans & spray, placards, painted, defaced, pictures & reads, wall, damage, started, portraits, spray \\
        \hline
    \end{tabular}
    \label{tab:discriminative_terms}
\end{table}

\begin{table*}[h!]
    \scriptsize
    \centering
    \caption{Gold and detected episodes (a maximum of five are included for brevity) for the ``2019 Hong Kong Legislative Protests'' key event. We specify the gold/detected episode attributes for each episode cluster in the following semicolon-separated format: core entity; action; object; time; location. ``Not detected'' denotes that no more episodes were generated by the model. We note the number of detected episodes beside the model name. We also color-code attributes which clearly align to a specific episode.}
    {\renewcommand{\arraystretch}{1.2}%
    \begin{tabular}{p{1.5cm}|p{2.5cm}|p{2.5cm}|p{2.5cm}|p{2.3cm}|p{2.2cm}}
        \hline
        \textbf{Model} & \textbf{Episode \#1} & \textbf{Episode \#2} & \textbf{Episode \#3} & \textbf{Episode \#4} & \textbf{Episode \#5} \\
        \hline
        \textbf{Gold}\newline(5 eps) &
        \cellcolor{CBBlue!15} Activists; headed; towards the Legislative Council Complex; 1 July 2019; Hong Kong & 
        
        \cellcolor{CBOrange!15} Protesters; stormed; the Legislative Council Complex; around 9:00 pm; Hong Kong; &
        \cellcolor{CBGreen!15} Protesters; damaged/defaced; portraits, furniture, emblem, etc.; 1 July 2019; Legislative Council Complex &
        
        \cellcolor{CBRed!15} Police; started using; tear gas to disperse protesters; 12:05 am 2 July; around the Legislative Council complex &
        
        \cellcolor{CBPurple!15}Police; arrested; individuals in connection with the incident; between 3 July and 5 July; Hong Kong\\
        \hline
        \textbf{K-means + \claude}\newline(4 eps) & Protesters; \textcolor{CBOrange}{storm} and \textcolor{CBGreen}{vandalize}; Legislative Council building; July 1, 2019; Legislative Council complex in Admiralty, Hong Kong & Hong Kong government; condemns; \textcolor{CBOrange}{protesters storming legislative building}; July 1, 2019; Hong Kong & Hong Kong protesters; express; demands for freedom and democracy; July 1, 2019; Hong Kong Legislative Council & \textcolor{CBRed}{Hong Kong police; adopt; more restrained tactics}; July 1, 2019; Hong Kong Legislative Council & Not detected \\
        \hline
        \textbf{EvMine + \claude}\newline(4 eps) & Protesters; \textcolor{CBGreen}{vandalize}; Hong Kong legislative building; July 1, 2019; Hong Kong legislative building & Protesters; \textcolor{CBGreen}{occupy and vandalize}; Hong Kong legislative chamber; July 1, 2019; Hong Kong legislative building & Protesters; \textcolor{CBGreen}{spray paint; slogans and demands}; July 1, 2019; Hong Kong legislative building & Protesters; \textcolor{CBGreen}{deface; Hong Kong emblem}; July 1, 2019; Hong Kong legislative building & Not detected \\
        \hline
        \textbf{Claude (\claude)}\newline(3 eps) & Protesters; \textcolor{CBOrange}{storm}; Hong Kong's Legislative Council; July 1, 2019; Hong Kong's Legislative Council building & Police; retreat and avoid confrontation; \textcolor{CBOrange}{protesters storming} Hong Kong's Legislative Council; July 1, 2019; Hong Kong's Legislative Council building & \textcolor{CBGreen}{Brian Leung Kai-ping; pulls off mask and reads protesters' demands; inside Hong Kong's Legislative Council}; July 1, 2019; \textcolor{CBGreen}{Legislative Council chamber} & Not detected & Not detected\\
         \hline
         \textbf{GPT-4}\newline(2 eps)& Hong Kong protesters; \textcolor{CBOrange}{storm} Legislative Council; government and police; July 1, 2019; Legislative Council Complex, Hong Kong & Hong Kong citizens; \textcolor{CBBlue}{march against extradition bill}; "Carrie Lam and Chinese government; June 2019; Various locations in Hong Kong & Not detected & Not detected \\
          \hline
         \textbf{EpiMine w/ \claude}\newline(7 eps) & Protesters; \textcolor{CBOrange}{broke into and occupied; Hong Kong's legislative building; July 1, 2019}; Hong Kong & Protesters; \textcolor{CBGreen}{vandalized}; the legislative building; \textcolor{CBGreen}{after breaking in}; Hong Kong & \textcolor{CBRed}{Police; fired tear gas at}; protesters; \textcolor{CBRed}{after midnight on July 1; outside the legislative building} & Carrie Lam; condemned; the protesters' actions; in a news conference at 4am on July 2; Hong Kong & \textcolor{CBPurple}{Police; began making arrests of; protesters involved; in the days after}; Hong Kong\\
    \end{tabular}
    }
    \label{tab:baselinecases}
\end{table*}

\subsection{Overall Results \& Analysis}
\par In Table \ref{tab:results}, EpiMine shows an average \textbf{80.8\%} increase in 5-precision, a \textbf{34.0\%} increase in 5-recall, and a \textbf{62.8\%} increase in 5-F1 over all baselines. Notably, despite both K-means and K-means + \claude \space being \textit{given the ground-truth number of episodes}, they are \textbf{significantly outperformed by EpiMine} (both the base model and no confidence ablation). Additionally, EvMine and EMiner, originally designed for key event and atomic action levels of event granularity, \textbf{fail to address the unique challenges of episode detection}. We further analyze our results through extensive quantitative and qualitative studies, including a detailed case study on the ``2019 Hong Kong Legislative Protest'' (as shown in Figure \ref{fig:event_hierarchy}), leading to the following takeaways:
\par{\textbf{1. LLMs require effective episode fragment clusters for synergistic episode estimation.}} As shown in Table \ref{tab:baselinecases}, LLMs without any initial clusters as guidance (\claude\space, GPT-4\footnote{https://chat.openai.com/}) fail to detect high-quality episodes, miss most ground-truth episodes, and include irrelevant atomic actions (e.g., ``Brian Leung pulls off mask''). Similarly, using low-quality baseline clusters results in poor performance. EvMine detects episodes that all reflect the same event, ``Protesters vandalized the Legislative Chamber''. While K-means produces more distinct episodes, it does not capture the most critical, gold episodes. In contrast, EpiMine's episodes are both distinct and meaningful, attributed to its cluster quality. This is quantitatively confirmed by  EpiMine-No LLM's competitive performance: using only EpiMine’s episode fragment clusters to compute episode representations---without any LLM summarization---still yields significantly better performance than all baselines (without or without LLM integration) on the Terrorism and Politics datasets, and remains highly competitive on Natural Disasters. This indicates the high quality of our fragment ranking and clustering.

\begin{figure}[!h]
    \centering
    \includegraphics[width=0.45\textwidth]{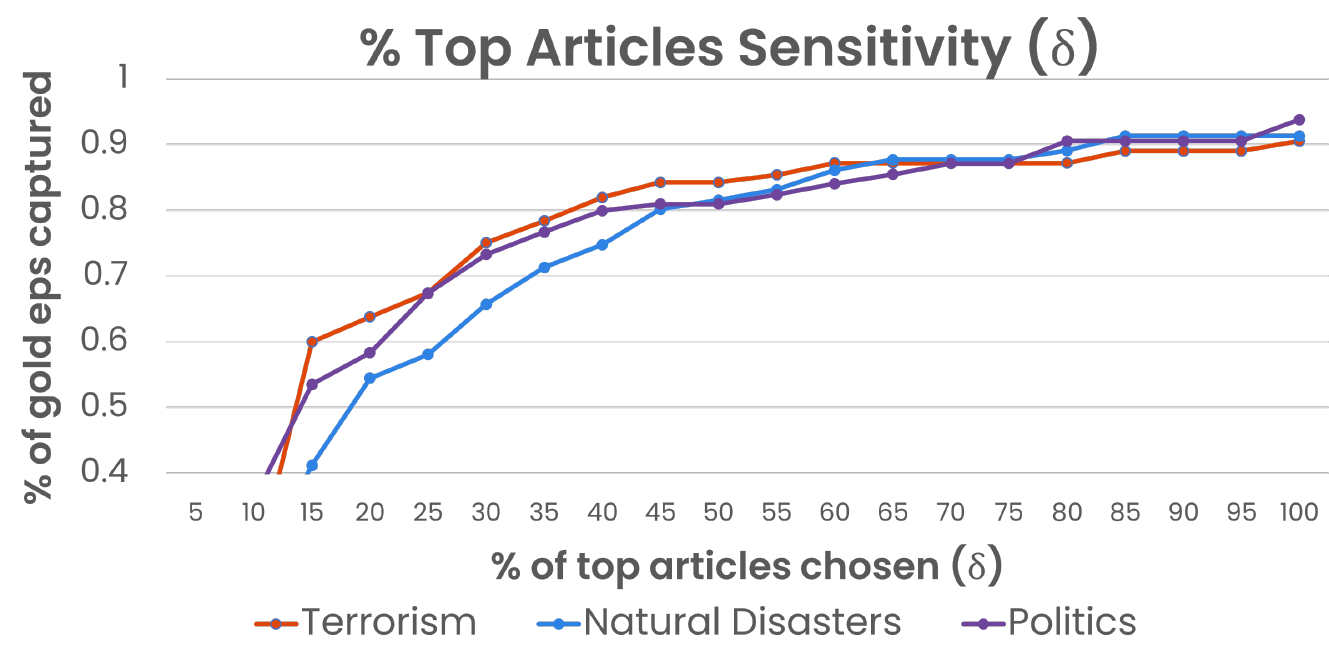}
    \caption{Percentage of key event's gold episodes captured in the $\delta$\% top articles chosen during the candidate episode estimation. Results averaged across themes.}
    \label{fig:top_episodes}
\end{figure}

\par EpiMine's clusters also elicit the LLM to identify more meaningful temporal information. Unlike most baseline episodes which have ``July 1, 2019'' as the time attribute, EpiMine's episodes feature more descriptive temporal cues: ``after breaking in'', ``after midnight'', ``in a news conference at 4 am on July 2''. Moreover, EpiMine's ``incorrect'' episode \#4 is a significant sub-event of the key event discussed by many articles. This \textit{strongly demonstrates the impact of EpiMine's candidate episode clusters} as input into the LLM; \textit{\textbf{LLMs alone cannot perform quality episode detection}}.

\par{\textbf{2. The strengths of discriminative co-occurrence complement those of cosine similarity.}} Table \ref{tab:discriminative_terms} illustrates the qualitative strengths of our novel discriminative co-occurrence metric. Both cosine similarity (CS) and discriminative co-occurrence (DC) offer different, complementary strengths. CS identifies similar words that play a \textit{similar role or are synonyms} within an episode (e.g., ``broke'', ``ransacked''), while DC identifies the \textit{key surrounding actions and objects} that co-occur within the same episode (e.g., ``slogans'', ``wall''). This is quantitatively supported by \textbf{\textit{TF-IDF}} and \textbf{\textit{No DC}} in Table \ref{tab:ablations}, which show a significant decrease in the quality of our top episode clusters without our salience and discriminative co-occurrence measures. We note that the politics dataset does show slight improvements in precision and F1 when discriminative co-occurrence is replaced, likely due to more term overlap across episodes (less distinct, sequential episodes).

\par{\textbf{3. Fragment ranking identifies top articles.}} In Figure \ref{fig:top_episodes}, we conduct a sensitivity analysis of the top articles chosen to estimate candidate episodes (Section \ref{sec:candidate_estimation}). We compare the gold episodes contained in the set of the top $\delta$\% articles as we vary $\delta$. By ranking the articles based on their likelihood of containing both high-quality and numerous episodes, we find that \textit{EpiMine's top article selection covers the vast majority of episodes} by $\delta = 25\%$ and more comprehensively around $\delta = 45\%$. This is significant as it helps \textit{minimize both the noise and the amount of data} needed to accurately detect all episodes. This is further supported by the EpiMine-Top ablation (Table \ref{tab:ablations}), which quantitatively shows that the top 25\% of the ranked fragments alone have competitive or higher performance than the baselines.

\par{\textbf{The Role of Confidence.}} The confidence metric influences EpiMine toward \textbf{more conservative} episode-segment classification by pruning segments with statistically insignificant confidence scores ($\leq \mu - \sigma$), which are unlikely to map to a single episode. Reducing this threshold increases the number of mapped segments, potentially improving recall by converting false negatives into true positives, but at the cost of reduced precision due to an increase in false positives. This trade-off is quantitatively demonstrated in the ``No-Confidence'' ablation (Table \ref{tab:results}), where omitting confidence leads to occasional gains in recall accompanied by declines in precision. Nevertheless, both configurations---with and without confidence---are significantly better than all baselines, allowing users to determine if including confidence aligns with their use case.

\section{Conclusion}
In this work, we proposed \textbf{EpiMine}, a novel, unsupervised episode detection method for large-scale news events. EpiMine performs (1) episode indicative term mining--- identifying combinations of salient terms that are likely to discriminatively co-occur \textit{within} an episode and not \textit{across} episodes, (2) episode partitioning, which partitions each article into approximate isolated episodes, (3) LLM-enhanced episode estimation, which clusters the top partitions into candidate episodes and synergizes with an LLM to produce fluent and meaningful episodes, and (4) episode-segment classification, which maps confident segments to their respective episode clusters. EpiMine significantly outperforms all baselines on the vast majority of key events, as shown through extensive quantitative and qualitative analysis.

\section{Limitations \& Future Work}
\par While EpiMine serves as an intuitive, unsupervised framework which demonstrates a more interpretable granularity for event analysis (episodes), it contains a few limitations that form the foundation for future, impactful research areas.

\par We note that the key event theme has an impact on EpiMine's performance. Specifically, natural disaster episodes are typically sequential and semantically distinct: disaster begins $\rightarrow$ warning $\rightarrow$ evacuation $\rightarrow$ damage/deaths $\rightarrow$ relief. As K-means is uniquely given $k$, the number of episodes, and relies on semantic similarity, it performs well with distinct episodes. However, we still see that its reliance on surface-level semantics leads to lower precision. Additionally, in the ablation studies shown in Table \ref{tab:ablations}, the politics dataset does show slight improvements in precision and F1 when \textit{discriminative co-occurrence is replaced}, due to more term overlap across episodes, resulting in less distinct, sequential episodes.

\par Further work towards the temporal analysis of episodes within articles can be explored, as well as extending our work to primarily multilingual news settings with low resources.

\section{Ethics Statement}
Based on our current methodology and results, we do not expect any significant ethical concerns, given that subtasks like episode detection within the news event extraction and analysis is a standard problem domain across data mining applications. Furthermore, having the method rely on zero supervision helps as a barrier to any user-inputted biases. However, one minor factor to take into account is any hidden biases that exist within the large language models used as a result of any potentially biased data that they were trained on. We used these pre-trained language models for refining the fluency of the detected episode clusters and did not observe any concerning results, as it is a low-risk consideration for the domains that we studied.

\section{Acknowledgements}
\par This work was supported by the National Science Foundation Graduate Research Fellowship. This research used the DeltaAI advanced computing and data resource, which is supported by the National Science Foundation (award OAC 2320345) and the State of Illinois. DeltaAI is a joint effort of the University of Illinois at Urbana-Champaign and its National Center for Supercomputing Applications.
\bibliography{custom}

\appendix

\section{Experimental Settings}
\label{sec:settings}
For implementing \textbf{EpiMine}, we use the following hyperparameters across all datasets: $\delta=25\%$, $sim\_thresh=0.75$. All other hyperparameters are set to their respective default values. We provide all experimental settings in Appendix \ref{sec:settings}.To determine statistical significance, we check for $\geq \mu - \sigma$. For our word representations, we use \texttt{bert-base-uncased}. For our sentence representations, we use \texttt{all-mpnet-base-v2}. We choose to use \texttt{Claude-2.1}\footnote{claude.ai/} for fluent candidate episode estimation due to its strong structured JSON/XML input and output formatting abilities. However, this proprietary model can be replaced with any open-source model as EpiMine is model-agnostic. We use only one NVIDIA GeForce GTX 1080 for all experiments; for non-API models, we utilize two NVIDIA-RTX A6000s.

\section{Baselines}
\label{sec:baselines}
We compare against the following methods using the evaluation metrics specified in Appendix \ref{sec:metrics}.

\begin{itemize}
    \item \textbf{K-means}~\cite{likas2003global}: No. of ground-truth episodes is given; clusters segments based on semantic similarity of ST ~\cite{reimers2019sentence} embeddings.
    \item \textbf{EvMine}~\cite{zhang2022unsupervised}: Unsupervised framework for key event detection that leverages peak phrases and detects communities using event-indicative features. We extend the original document-level method to the segment level for episode detection.
    \item \textbf{EMiner}~\cite{jiao2023unsupervised}: Unsupervised event chain mining that performs atomic action clustering. For episode detection, we map its final output, a list of events, back to the original sentences from which each event was extracted, treating these sentences as segments. To retrieve more episode-associated segments, we use ST \cite{reimers2019sentence} to select the $k$ most similar segments to each cluster sentence.
\end{itemize}
We also include the following full and partial ablations of EpiMine (clusters segments from all articles vs. top $\delta$\% articles, respectively):
\begin{itemize}
    \item \textbf{No Confidence}: A full ablation, where all input segments are classified based on the episode with max cosine similarity instead of using the confidence score from Equation \ref{tag:episode_confidence}.
    \item \textbf{No LLM}: We take the estimated episode clusters from Section \ref{sec:candidate_estimation} that normally would have been inputted into the LLM, and instead use them to compute our episode representations directly. These representations are used for our classification step (Section \ref{sec:classification}), run on the full dataset with confidence.
    \item \textbf{EpiMine-Top}: A partial ablation which directly outputs the intermediate episode clusters formed based on the top articles identified in Section \ref{sec:candidate_estimation} without inputting them into the LLM-based episode estimation step.
    \item \textbf{TF-IDF}: A partial ablation which replaces the salience and synonym expansion step (Section \ref{sec:term_mining}) with TF-IDF.
    \item \textbf{No DC}: A partial ablation which replaces the discriminative co-occurrence score (Equation \ref{tag:discriminative}) with raw pair frequency.
\end{itemize}

\section{Evaluation Metrics}
\label{sec:metrics}
\par Following a recent work on key event detection \cite{zhang2022unsupervised}, we adapt the $k$-prec, $k$-recall, and $k$-F1 metrics to quantitatively evaluate episode detection performance--- specifically, how the model's top-$k$ segments within each detected episode align with the ground truth episodes.

\par Formally, suppose there are $N$ ground truth episodes $\mathcal{G} = \{G_1, G_2, \dots, G_N\}$, each of which is a set of text segments related to its corresponding episode. $\mathcal{E} = \{E_1, E_2, \dots, E_K\}$ are the model predicted episodes, each of which is a ranked list of segments, and $E_{j, k}$ means the top-$k$ segments within $E_j$. Then, the $k$-metrics are defined as follows:
\begin{equation*}
\small
    \begin{split}
        \text{\textbf{k-prec}} &= \frac{\sum_{G_i \in \mathcal{G}} \mathds{1}(\exists E_j \in \mathcal{E}, E_{j,k} \cap G_i \ge \frac{k}{2})}{\sum_{E_j \in \mathcal{E}} \mathds{1}(|E_j| \ge k)} \\
    \text{\textbf{k-recall}} &= \frac{\sum_{G_i \in \mathcal{G}} \mathds{1}(\exists E_j \in \mathcal{E}, E_{j,k} \cap G_i \ge \frac{k}{2})}{N}\\
    \text{\textbf{k-F1}} &= \frac{2 \cdot \text{k-prec} \cdot \text{k-recall}}{\text{k-prec} + \text{k-recall}}\\
    \end{split}
\end{equation*}

\section{Key Event Corpus Dataset Construction \& Annotation}
\label{sec:keyeventinfo}

\par Given that our task is novel and no large-scale key event-specific news corpus is available for this task where the key events are guaranteed to contain distinguishable episodes, we briefly discuss how we collect the input corpus from online news data. Given our set of key events (as listed in Section \ref{keyeventlist}), we first scrape the external reference list from their corresponding Wikipedia page and select the news articles that have been published within two months of given key event's start date (e.g., all articles selected for ``January 6 2021 Capitol Attack'' would have been published between November 6-March 6). This is important as we want to prioritize the news articles which focus on describing the episodes of the key event and their corresponding aspects as opposed to primarily opinions or analyses. This allows us to motivate our task as one critical for currently evolving key events which required a more fine-grained episodic timeline. Furthermore, it is consequently \textit{unlikely} for a single article to cover \textit{all the episodes} and \textit{exclusively} episodes under a key event. Despite this being more challenging, it is acceptable as the goal of our task is to extract only the key event-related episodes, which must be substantiated by multiple documents in either case.
\par During the collection process, we targeted selecting a diverse set of key events topics within a theme. For instance, we attempted to cover every type of ``natural disaster'', including tornados, wildfires, and etc. When selecting key events, we leave out those with less than $20$ hyperlinks in the Wikipage and manually inspect at least 20 articles per event in order to ensure quality. Table~\ref{table:dataset_stats} summarizes the statistics for these datasets. We also construct a background news corpus of approximately 4,000 long news articles using the New York Times corpus for topic categorization \cite{meng2020discriminative}. 

\par For the annotation process, we had each of the four individual annotators (computer science graduate students) manually identify a ground-truth description for each of the episodes under every key event. The descriptions and a one-shot annotation demonstration (assigning either an episode ID per segment, or `M'/`X' if it describes multiple or none) are provided in our automatic annotation prompt (Appendix \ref{sec:annotation_prompt}).

\par The same annotators each manually annotated a subset of segments (25 articles, 300 segments in total) with either their corresponding episode ID or `-1' if the segment was either `X' and `M'. The human-automatic annotation agreement is shown in Table \ref{tab:agreement} with three versions of intra-class correlation (ICC) and Cohen’s $\kappa$). The Cohen's $\kappa$ indicates substantial agreement, and the all three versions of ICC indicate good reliability.

\begin{table}[h!]
    \small
    \centering
    \begin{tabular}{l|c|c|c|c}
    \midrule
         & \textbf{Cohen's $\kappa$} & \textbf{ICC1k} & \textbf{ICC2k} & \textbf{ICC3k} \\
         \midrule
        \textbf{Score} & 0.614 & 0.772 & 0.772 & 0.773 \\
        \midrule
    \end{tabular}
    \caption{Agreement scores between the human and LLM annotation of each article segment.}
    \label{tab:agreement}
\end{table}

\section{Identifying Salient Terms for Episode Detection}
\label{sec:salient}

\par We define the salience score of a term $w_i$ within segment $s$ as the following function, where $freq(w_i)$ is the number of key event segments that $w_i$ is contained in, $N_{bg}$ is the number of news articles in the background corpus we construct (using general New York Times articles), and $bgf(w_i)$ is the number of background articles that $w_i$ is present in. 
\begin{equation}
    \label{tag:salience}
    \begin{split}
        \textbf{Salience}(w_i) = &\left(1 + \log^2\left(freq(w_i)\right)\right) \\
        & \times \log\left(\frac{N_{bg}}{bgf(w_i)}\right)
    \end{split}
\end{equation}

\par Stop words and infrequent terms ($freq(w_i) < 5$) are assigned a salience score of $-1$. A key event's set of salient terms $T$ is comprised of the terms with a salience score \textit{above the mean salience} across the entire vocabulary. In the case of infrequent synonyms used by a journalist as a stylistic choice (e.g., ``demonstrations'', ``marches''), we expand $T$ with terms that are similar (cosine-similarity) to their static word representations (average of its contextualized word embeddings across entire key event corpus).

\section{Key Event News Article Pre-Processing}
\label{sec:segment}
\par Given that the expected output for the episode detection task is a \textit{cluster of text segments}, we first must segment each key event news article. We would like to ideally preserve both the primary aspects (e.g., core entities and their actions) and peripheral aspects (e.g., reactions to a core entity's action) relevant to that episode, which may be helpful for cross-document episode co-reference resolution. In order to do this, we utilize the text segmentation method, C99 \cite{choi2000advances}. Furthermore, in order to assist with the cohesiveness of the segment, we employ entity co-reference resolution before performing segmentation, which assists with retaining the context across text segments (``They surrounded the legislative building [...]'' $\rightarrow$ ``The protesters surrounded the legislative building [...]''). Our core methodology is given these text segments (in their raw form, without co-references resolved) and their source articles as the primary inputs.

\section{Additional Details for Episode Partitioning}
\label{sec:partitioning_details}
\par We note that for determining semantic similarity between the terms of two segments, we use both (1) the average cosine similarity between all unordered pairs of terms between segment (i-1) and (i), and (2) the cosine similarity between the average of static term representations in $(i-1)$ and the average of static term representations in $(i)$. Furthermore, we filter out any non-salient segments before episode partitioning to avoid any influence of noisy segments (e.g., journalist's analysis, summary statements, historical comparisons, and other generic noise) on the quality of our episode fragments. 

\par Finally, following \cite{wang-etal-2021-x, kargupta-etal-2023-megclass}, we take the harmonic mean of all pairwise discriminative co-occurrence scores instead of a simple average. This allows us to prioritize the more salient \textit{and} discriminative terms when determining the episode partitions. For instance, if ``protesters'' consistently occurs throughout the majority of episodes and thus has a low average discriminative co-occurrence, then it is not as informative for episode partitioning.

\section{Claude-2 Prompt \& Example for Candidate Episode Estimation}
\label{sec:prompt}

\textbf{Prompt.} We use the following prompt for estimating fluent candidate episodes from our input episode fragment clusters. We denote $k$ as the number of episode fragment clusters outputted after clustering the top article episode fragments in Section \ref{sec:candidate_estimation}.

\texttt{\textbf{Task:} You are a key news event analyzer that is aiming to detect episodes (a representative subevent that reflects a critical sequence of actions performed by a subject at a certain and/or location) based on text segments from different news articles. Given the above groups of article segments, predict at least 2 and at most \{k\} potential episodes of the key event. Some groups may fall under the same episode. Output your answer inside the tags <answer></answer> as a JSON object where each item is also a JSON with the key "title" with the value containing the [subject, action, object, time, location] of the episode, a key "keywords" with the string value being a list of 5-10 associated keywords unique to that specific episode, and a final key "example\_sentences" with a value being a list of 2-5 extracted sentences from the input segment groups. Feel free to output less than \{k\} episodes if you feel that any are redundant (could fit under an existing candidate episode). The title, keywords, and example sentences of a predicted episode should not be able to be placed under another different predicted episode.}\\

\noindent\textbf{Example.} Below, we provide an example of EpiMine's candidate episode estimation step (Section \ref{sec:candidate_estimation}). Specifically, the LLM identifies the core attributes (subject, action, object, time, location) of each unique cluster, relevant keywords, and top extracted text segments given the input clusters:

\texttt{\\'\textbf{title}': ['Protesters', 'storm and vandalize', 'Hong Kong's Legislative Council building', 'July 1, 2019', 'Legislative Council building in Admiralty, Hong Kong']} \\
\texttt{'\textbf{keywords}': ['vandalism, graffiti, violence, escalation, ransacking']} \\
\texttt{'\textbf{example\_sentences}': ['Hundreds of anti-extradition bill protesters finally broke into the legislature after many hours of attacking the public entrance and ransacked the building, including displaying the colonial Hong Kong flag in the chamber.', 'Slogans on the wall read: ``Murderous regime'', and ``There are no rioters only a tyrannical regime.'']}

\section{Claude-2 Prompt for Dataset Annotation}
\label{sec:annotation_prompt}

We automatically annotate our dataset using Claude-2.1 using the prompt below (before an additional human-verification stage):
\newline\newline
\texttt{You are a news event analyzer that labels text segments of a news article with their matching event episode description. I will give you several text segments, and several episodes of a key event in tuples. We define an episode as the following: an episode is a set of thematically coherent text segments discussing a particular set of core entities performing actions for or towards an object(s) at a certain time and/or location during a real-world key event. The entities, actions, objects, time, and location can all be considered aspects of an episode.
\newline\newline
[one-shot demonstration \& format specification]
\newline
Please help classify the text segments under different episodes (the output value for each segment should be an integer key of each episode). If you think a text segment cannot be used to describe any episodes, please use "X" in the output to indicate the lack of an episode tuple number for that segment. If a text segment is very general, does not describe the key event at hand, or can be matched to multiple episodes, then please use a "M" in the output to indicate the multiple episode mapping for that segment. There should be a value assigned to each of the {len(segments)} segments (segment\_0, ..., segment\_{len(segments)-1}).
}

\end{document}